\documentclass[runningheads]{llncs}

\usepackage{eccv}

\usepackage{eccvabbrv}

\usepackage{graphicx}
\usepackage{booktabs}

\usepackage{bm}
\usepackage{bbding}
\usepackage{colortbl}
\usepackage[misc]{ifsym}
\usepackage{multirow}

\usepackage[accsupp]{axessibility}  

\usepackage{hyperref}

\usepackage{orcidlink}

\makeatletter
\newcommand{\thickhline}{%
    \noalign {\ifnum 0=`}\fi \hrule height 1pt
    \futurelet \reserved@a \@xhline
}
\makeatother
\newcommand{\pub}[1]{\color{gray}{\tiny{[{#1}]}}}

\let\titleold\title
\renewcommand{\title}[1]{\titleold{#1}\newcommand{\thetitle}{#1}}
\def\maketitlesupplementary
   {
   \newpage
    {
    \centering
    \Large
    \textbf{\thetitle}\\
    \vspace{0.5em}Supplementary Material \\
    \vspace{1.0em}
    }
   }

\begin{document}

\title{Controllable Navigation Instruction Generation with Chain of Thought Prompting} 

\titlerunning{C-Instructor}

\author{Xianghao Kong\inst{1}\orcidlink{0009-0004-9865-4105}\thanks{Equal contribution. \ \textsuperscript{\Letter} Corresponding author.} \and
Jinyu Chen\inst{1}\orcidlink{0009-0002-7106-8312}$^\star$ \and
Wenguan Wang\inst{2}\orcidlink{0000-0002-0802-9567}\textsuperscript{\Letter} \and
Hang Su\inst{3}\orcidlink{0000-0001-8294-6315} \and \\
Xiaolin Hu\inst{3}\orcidlink{0000-0002-4907-7354} \and
Yi Yang\inst{2}\orcidlink{0000-0002-0512-880X} \and
Si Liu\inst{1}\orcidlink{0000-0002-9180-2935}\textsuperscript{\Letter}}

\authorrunning{X.~Kong et al.}

\institute{School of Artificial Intelligence, Beihang University\and
College of Computer Science and Technology, Zhejiang University\and
Department of Computer Science and Technology, Tsinghua University\\
\url{https://github.com/refkxh/C-Instructor}
}

\maketitle

\begin{abstract}
Instruction generation is a vital and multidisciplinary research area with broad applications. 
Existing instruction generation models are limited to generating instructions in a single style from a particular dataset, and the style and content of generated instructions cannot be controlled. 
Moreover, most existing instruction generation methods also disregard the spatial modeling of the navigation environment. 
Leveraging the capabilities of Large Language Models (LLMs), we propose \textsc{C-Instructor}, which utilizes the chain-of-thought-style prompt for style-controllable and content-controllable instruction generation.
Firstly, we propose a Chain of Thought with Landmarks (CoTL) mechanism, which guides the LLM to identify key landmarks and then generate complete instructions. CoTL renders generated instructions more accessible to follow and offers greater controllability over the manipulation of landmark objects. 
Furthermore, we present a Spatial Topology Modeling Task to facilitate the understanding of the spatial structure of the environment. 
Finally, we introduce a Style-Mixed Training policy, harnessing the prior knowledge of LLMs to enable style control for instruction generation based on different prompts within a single model instance. 
Extensive experiments demonstrate that instructions generated by \textsc{C-Instructor} outperform those generated by previous methods in text metrics, navigation guidance evaluation, and user studies. 
  \keywords{Instruction generation \and Vision-and-language navigation}
\end{abstract}

\section{Introduction}
\label{sec:intro}
\begin{figure}
    \centering
    \includegraphics[width=\linewidth]{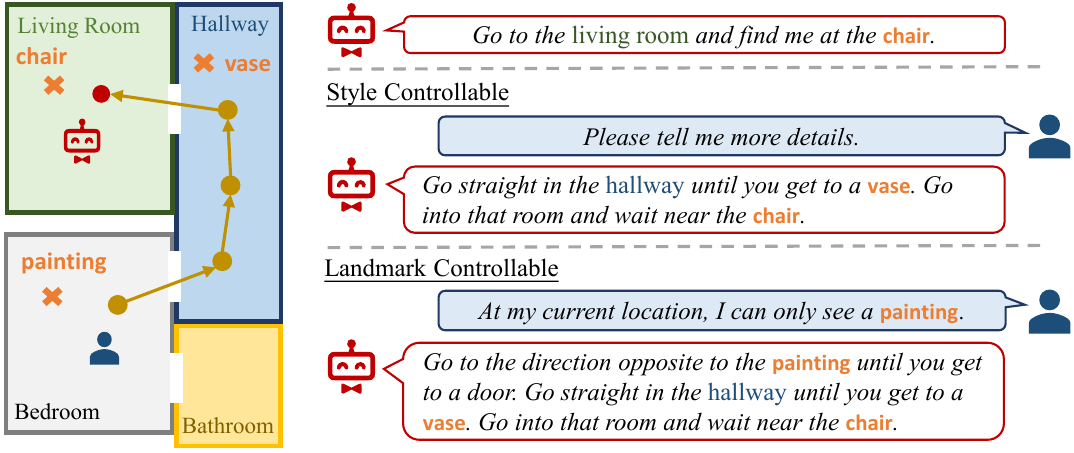}
      \caption{\textsc{C-Instructor} possesses the ability to control the linguistic style of generated instructions, and the ability to manipulate landmarks within the instructions (\S\ref{sec:intro}). 
      }
   \label{fig:fig1}
\end{figure}
Developing an agent capable of communicating with humans in natural language and accomplishing specific tasks in its environment is a crucial goal for researchers in the field of embodied AI. Such an agent needs two key abilities: the first one is to execute specific tasks based on human instructions, and the second one is to provide interactive feedback and guidance to humans based on environmental information. Regarding the first ability, one of the most typical tasks is vision-and-language navigation (VLN)~\cite{anderson2018vision}, which has garnered extensive research interest~\cite{qi2020reverie,zhu2021soon,ku2020room,thomason2020vision,2019Help,2020ALFRED,mehta2020retouchdown,huang2022assister} and developed fast in recent years~\cite{gao2021room, chen2022reinforced, zhao2022target, gao2023adaptive, gao2023room, wang2020active, wang2021structured, liu2023bird, wang2023dreamwalker, liu2024volumetric, an2024etpnav, an2023bevbert, wang2023gridmm, qiao2023march, he2024frequency, li2024panogen, yang2024doraemongpt}. 

Regarding the implementation of the second capability, \ie, machine feedback, one of its prominent facets, instruction generation, has been a long-standing area of multidisciplinary research dating back to the 1960s~\cite{lynch1964image}.
The instruction generation model can be used for describing a path explored by a robot to a human in human-robot collaboration tasks. In practical scenarios, it can be applied to intelligent guidance for the visually impaired~\cite{huang2022assister}, foster human-machine trust~\cite{wang2023lana}, and provide guidance in hazardous scenarios, \etc.
An instruction generation model fulfilling the prerequisites of human-machine collaboration must possess the following two capabilities~\cite{qi2020reverie,ku2020room}, \ie, executability and controllability. For executability, instructions are supposed to exhibit high linguistic quality and provide clear guidance at navigational junctions. 
For controllability, control over instruction generation in style and content is also of essential importance to improve communication efficiency. For example, when the instruction recipient is acquainted with the environment, it is more efficient to generate instructions with higher levels of abstraction. Additionally, the guidance provided in the instructions may need adjustments based on the landmarks that the instruction recipient focuses on in the environment. 

To enhance the executability and controllability of instruction generation models, we propose a \underline{C}ontrollable Navigation \underline{Instructor}  (\textsc{C-Instructor}), which possesses the ability to generate easily executable instructions with high linguistic quality, as well as the capability to controllably generate instructions in various linguistic styles with different landmarks (\cref{fig:fig1}). 
\textsc{C-Instructor} primarily encompasses the following four technological contributions: 
\textbf{First}, to enhance the linguistic quality of instruction generation and handle different styles of instructions neatly, we propose an adapter structure that effectively incorporates path information into the GPT-based Large Language Model (LLM)~\cite{gao2023llama}.
\textbf{Second}, to improve the executability of generated instructions, we present a training strategy involving a Chain-of-Thought with Landmarks (CoTL) mechanism and a Spatial Topology Modeling Task (STMT). CoTL employs a step-by-step thinking~\cite{wei2022chain} approach to guide the model to identify crucial landmarks before generating complete instructions; 
STMT incorporates spatial connectivity prediction as an auxiliary task in training to facilitate the understanding of the topological structure of the environment. 
\textbf{Third}, in order to generate instructions in various styles with a single model instance, we introduce a Style-Mixed Training (SMT) policy, in which different styles of instructions are jointly learned. Distinct instruction styles are trained using prompts as differentiation, enabling control over the style of generated instructions.
\textbf{Fourth}, the collaboration between CoTL and SMT enhances the capabilities of crucial navigation waypoints localization and spatial direction guiding, thus improving the executability of the generated instructions. 
Benefiting from SMT and CoTL, \textsc{C-Instructor} allows control over the generation style of instructions and attention to specific objectives while maintaining high linguistic quality of generated instructions.

In our experiments, \textsc{C-Instructor} significantly outperforms previous instruction generation methods~\cite{Fried_Hu_Cirik_Rohrbach_Andreas_Morency_Berg-Kirkpatrick_Saenko_Klein_Darrell_2018,tan2019learning,wang2023lana,wang2022counterfactual} across different linguistic metrics on four indoor/outdoor benchmarks~\cite{anderson2018vision,qi2020reverie,huang2022assister,ku2020room}. In addition, it proves to be an effective means of data augmentation for VLN training over previous speaker models~\cite{Fried_Hu_Cirik_Rohrbach_Andreas_Morency_Berg-Kirkpatrick_Saenko_Klein_Darrell_2018,tan2019learning,wang2022counterfactual,wang2023lana}. Moreover, instructions generated by \textsc{C-Instructor} demonstrate enhanced navigation guidance capabilities in both instruction following model experiments and human evaluations. 

\section{Related Work}
\label{sec:related}


\noindent\textbf{Navigation Instruction Generation.} 
The study of generating linguistic instruction for navigation can date back to Lynch’s work~\cite{lynch1964image} in the 1960s.
Early efforts~\cite{Shawn1986turn,allen1997knowledge} investigated the human cognitive mechanism for describing routes. They found that navigation direction is associated with the cognitive map~\cite{kuipers1978modeling} and influenced by various factors including cultural background~\cite{vanetti1988communicating} and genders~\cite{hund2006getting}. This area has long been overlooked by the computer vision academia and is simply viewed as a data augmentation tool for VLN. However, it holds significant practical relevance, \eg, establishing human-machine trust~\cite{wang2023lana} and facilitating blind navigation~\cite{huang2022assister}.
Fried \textit{et al} \cite{Fried_Hu_Cirik_Rohrbach_Andreas_Morency_Berg-Kirkpatrick_Saenko_Klein_Darrell_2018} first proposed a LSTM-based instruction generation model to augment training samples and re-weight the route choice of the navigator. There are three primary aspects for the advancement of instruction generation: elevated linguistic quality, finer-grained directives, and longer, more intricate instructions.
In order to enhance the quality of instructions, some methods introduce supplementary information like external knowledge~\cite{Zeng_Wang_Wang_Yang_2023} and landmark information~\cite{wang2022less,zhang2023vln}, build instruction template~\cite{zhang2023vln} and utilize larger language models~\cite{wang2022less}.
\cite{kamath2023new,zhang2023vln,he2021landmark,zhu2020babywalk,hong2020sub} generate fine-grained alignment between language and navigation paths. To build more intricate instructions, \cite{Jain_Magalhaes_Ku_Vaswani_Ie_Baldridge_2019,zhu2020babywalk,Liu_2021_ICCV} cross-connect paths to generate longer instruction-trajectory pairs. 
Methods like~\cite{wang2023lana,wang2022counterfactual,dou2022foam} also consider instruction generation and following as dual tasks, and employ joint-optimization or cycle-consistent learning to promote navigation performance and instruction generation quality.

Previous deep-learning-based methods~\cite{Fried_Hu_Cirik_Rohrbach_Andreas_Morency_Berg-Kirkpatrick_Saenko_Klein_Darrell_2018,tan2019learning,wang2022counterfactual,wang2023lana} can only generate navigation instructions in a single style with limited linguistic quality and no controllability. By leveraging LLMs, \textsc{C-Instructor} notably enhances the linguistic quality of instructions. Moreover, \textsc{C-Instructor} provides style and content controllability in a single model instance via SMT and CoTL respectively.


\noindent\textbf{Parameter-Efficient Fine-Tuning.} 
The pre-training and fine-tuning paradigm has demonstrated remarkable efficacy in VLN and various other tasks. However, as model parameters grow exponentially and downstream task data remain limited, full-scale fine-tuning fails to yield robust performance on downstream tasks due to overfitting and catastrophic forgetting. The approach known as Parameter-efficient Fine-tuning (PEFT), involving the selective freezing of a significant portion of the model's parameters while training only a small subset, has met success in numerous domains. 
PEFT has proven highly effective in adapting pre-trained models like CLIP~\cite{radford2021learning}, BERT~\cite{devlin2018bert}, and GPT~\cite{brown2020language,touvron2023llama} to downstream tasks.
There are three main types of PEFT methods, namely prefix finetuning, reparameterization, and adapter. Prefix finetuning methods like~\cite{li2021prefix,Zhou_Yang_Loy_Liu_2022,lester2021power,liu2023gpt} feed learnable prompts into the model to learn task-specific knowledge. The methods~\cite{hu2021lora,karimi2021compacter} use reparameterization to reduce the amount of trainable parameters. 
Approaches employing adapters \cite{gao2023llama,zhang2023llama} adeptly accommodate inputs from diverse modalities and various downstream tasks by incorporating additional layers into the pre-trained network.

Understanding the spatial topology of the navigation environment is essential for the instruction generator to guide the instruction follower. Based on adapter PEFT methods~\cite{gao2023llama,zhang2023llama}, \textsc{C-Instructor} introduces a trajectory encoder to incorporate spatial information into the LLM. Moreover, \textsc{C-Instructor} includes STMT to facilitate the understanding of spatial connectivity of the environment.

\section{Methodology}
\label{sec:method}
\subsection{Task Formulation}
\label{sec:formulation}
The instruction generation model is required to generate the instruction $\textit{X}\!=\!\{\bm{x}_1,\bm{x}_2,..., \bm{x}_S\}$ with $S$ words that provides guidance for following the given path $\textit{R}\!=\!\{r_1,r_2,..., r_T\}$ with $T$ steps.
At a given time step $t$, $r_t$ is composed of the panoramic observation $o_t$ and action $a_t$.
The objective of model parameters $\bm{\theta}$ is to maximize the likelihood of the target instruction $\textit{X}^{ *}$:
\begin{equation}
    \bm{\theta}^*=\mathop{\arg\max}\limits_{\bm{\theta}} \log p (\textit{X}^{ *} | \textit{R},\bm{\theta}).
\end{equation}

\begin{figure}
  \centering
  \begin{subfigure}{0.515\linewidth}
    \centering
    \includegraphics[width=\linewidth]{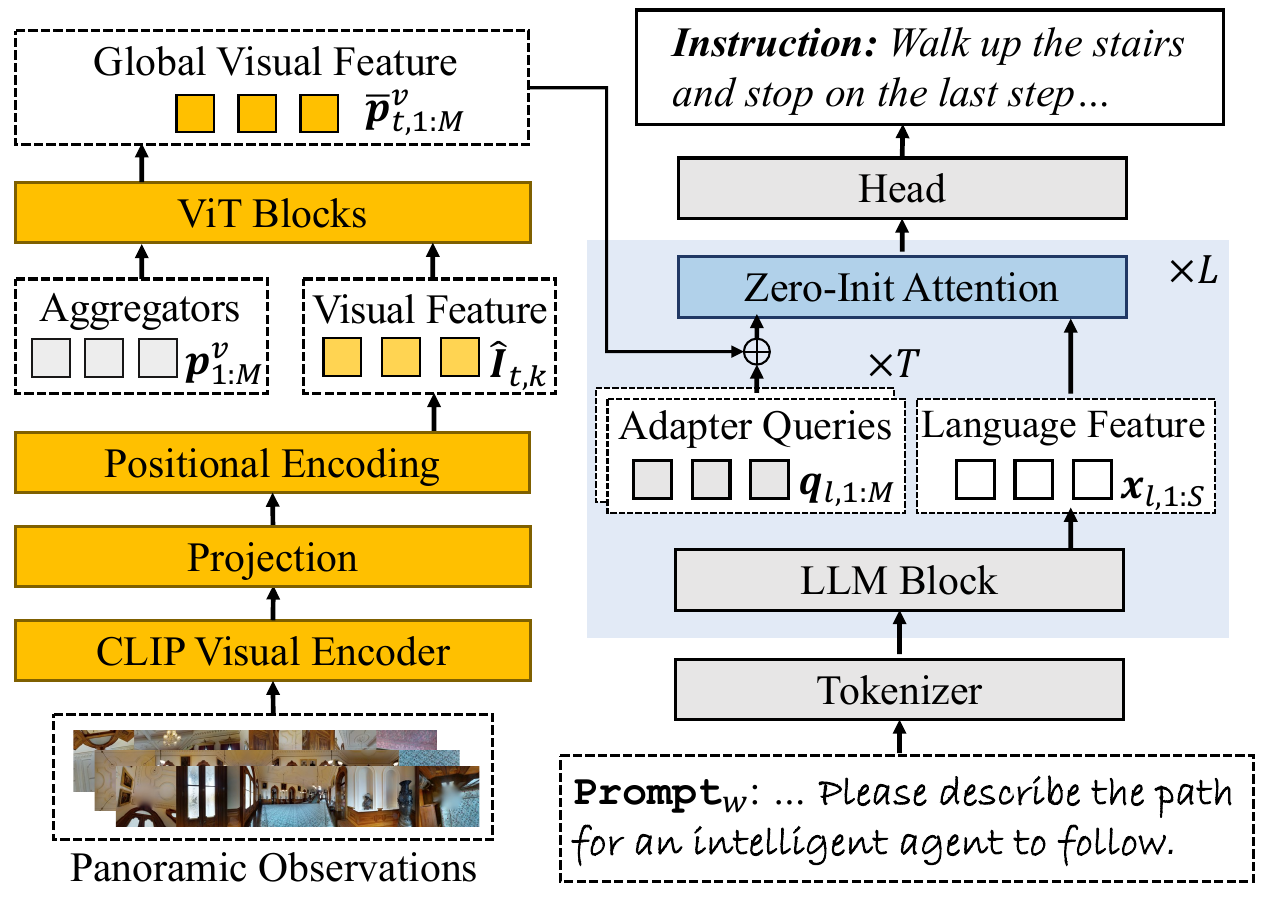}
    \caption{The overall framework of \textsc{C-Instructor} (\S\ref{sec:framework}) including Trajectory Encoder and LLM Adapter.}
    \label{fig:framework}
  \end{subfigure}
  \hfill
  \begin{subfigure}{0.47\linewidth}
    \centering
    \includegraphics[width=\linewidth]{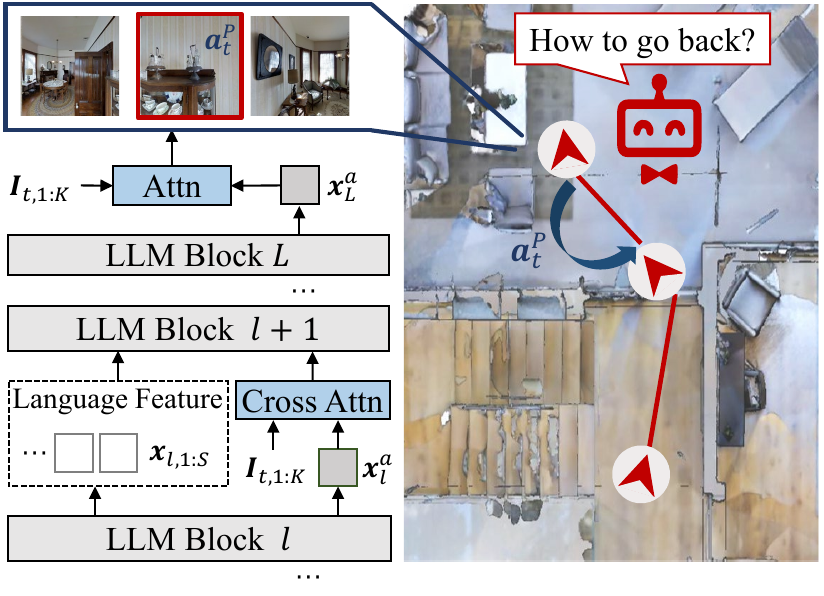}
    \caption{Details of STMT (\S\ref{sec:stmt}). In STMT, \textsc{C-Instructor} selects backtracking action that leads back to previous viewpoint.} 
    \label{fig:stmt}
  \end{subfigure}
  \caption{Overall framework of \textsc{C-Instructor} (\S\ref{sec:framework}) and details of STMT (\S\ref{sec:stmt}).}
  \label{fig:framework_and_stmt}
\end{figure}


\subsection{Overall Framework}
\label{sec:framework}
To leverage the linguistic capabilities of LLMs, we employ an adapter-based~\cite{gao2023llama} approach in \textsc{C-Instructor} to embed actions and visual observations. 
The adapter consists of two components: the Trajectory Encoder and the LLM Adapter. The overall structure is shown in \cref{fig:framework}.

\noindent\textbf{Trajectory Encoder.} 
The trajectory encoder encodes the viewpoint and action information for each step along the path into visual features. In the Matterport3D Simulator~\cite{2018Vision}, a panoramic observation $o_t$ at time step $t$ is partitioned into $K\!=\!36$ subview images $\{v_{t,k}\}_{k=1}^{K}$, where the action $a_t$ is represented using the index of the subview image corresponding to the motion direction. 
First, we extract visual features for each subview image using the CLIP~\cite{radford2021learning} visual encoder followed by a linear projection layer with Layer Normalization~\cite{ba2016layer}:
\begin{equation}
    \bm{I}_{t,k} = \texttt{layer\_norm}(\texttt{linear}(f_{CLIP}(v_{t,k}))),
\end{equation}
where $\bm{I}_{t,k}\!\in\!\mathbb{R}^{1\times D_I}$, $v_{t,k}\!\in\!\mathbb{R}^{224\times 224\times 3}$. 
To distinguish the spatial and temporal relation of each view, we add a spatial positional encoding $pos^v_k$ and a history encoding $pos^h_t$ to $\bm{I}_{t,k}$. 
To represent action information, we introduced a special token $pos^a$ for the action view $a_t$ and another token $pos^o$ for non-action views: 
\begin{equation}
 \hat{\bm{I}}_{t,k} = 
\left\{
    \begin{aligned}
    & \bm{I}_{t,k}+ pos^v_k + pos^h_t + pos^a,~~~~\text{if $k = a_t$}\\
    & \bm{I}_{t,k} + pos^v_k + pos^h_t + pos^o,~~~~\text{otherwise}.
    \end{aligned}
    \right.
\end{equation}
Subsequently, we concatenate $M$ aggregator tokens $\bm{p}^v_{1:M}$ with $\hat{\bm{I}}_{t,1:K}$ along the length dimension and then feed them into several ViT~\cite{dosovitskiy2020image} blocks to aggregate global features for step $t$:
\begin{equation}
   [\overline{\bm{p}}^v_{t,1:M};\overline{\bm{I}}_{t,1:K}] =f_{ViT}([\bm{p}^v_{1:M};\hat{\bm{I}}_{t,1:K}]),
\end{equation}
where $\bm{p}^v_{1:M}\!\in\! \mathbb{R}^{M \times D_p}$; $\overline{\bm{p}}^v_{t,1:M}$ is the trajectory feature representation at step $t$.

\noindent\textbf{LLM Adapter.} We introduce the trajectory features into LLM via layer-wise adapting. We utilize $\texttt{adapter}_l(\cdot,\cdot)$ to integrate the trajectory features $\overline{\bm{p}}^v_{t,1:M}$ into $\bm{x}_{l,1:S}$, which is the output of $l$-th LLM transformer block:
\begin{equation}
    \widetilde{\bm{x}}_{l,1:S} = \text{\texttt{adapter}}_l(\overline{\bm{p}}^v_{t,1:M},\bm{x}_{l,1:S}).
\end{equation}
Here $\widetilde{\bm{x}}_{l,1:S}$ replaces the $\bm{x}_{l,1:S}$ in the subsequent LLM blocks. Next, we will detail the structure of $\texttt{adapter}_l(\cdot,\cdot)$. We add the trajectory features $\overline{\bm{p}}^v_{t,1:M}$ with the $l$th-layer's adapter query $\bm{q}_{l,1:M}$ and map them to the textual space through a linear layer $\text{\texttt{linear}}_l(\cdot)$:
\begin{equation}
    \widetilde{\bm{p}}_{l,t,1:M} = \text{\texttt{linear}}_l (\overline{\bm{p}}^v_{t,1:M}+ \bm{q}_{l,1:M}).
\end{equation}
Next, we concatenate the $\{\widetilde{\bm{p}}_{l,t,1:M}\}_{t=1}^T$ in the order of $t$:
\begin{equation}
    \bm{\rho}_{l,1:V} = \text{\texttt{concat}}(\{\widetilde{\bm{p}}_{l,t,1:M}\}_{t=1}^T),~~V\!=\!T\!\times\!M .
\end{equation}
To preserve the natural language capabilities of the LLM, we use zero-initialized attention~\cite{zhang2023llama} to get $\widetilde{\bm{x}}_{l,1:S}$:
 \begin{equation}
     \widetilde{\bm{x}}_{l,1:S} = \text{\texttt{zero\_attn}}( [\bm{\rho}_{l,1:V};\bm{x}_{l,1:S}] ).
 \end{equation}
Based on this model structure, we design STMT (\S\ref{sec:stmt}) to improve the model's spatial awareness, and CoTL (\S\ref{sec:cotl}) to enhance the model's perception of landmarks. Finally, through SMT (\S\ref{sec:smt}), we achieve style-controlled instruction generation. In subsequent sections, we utilize $[R; W]$ to denote the model's input, where $R$ represents the path input, and $W$ stands for the language input.

\subsection{Spatial Topology Modeling Task (STMT)}
\label{sec:stmt}
Understanding the spatial relationships between different viewpoints is fundamental for generating navigation instructions. 
LLMs and visual encoders are typically trained on data from the Internet with few embodied-type data. Consequently, they possess limited spatial cognition abilities. Therefore, we introduce STMT as an auxiliary task to enhance the model's spatial perception capability.


In STMT, the model predicts actions between adjacent viewpoints along a trajectory. As the actions along the navigation path are already represented through location encoding, we make the model predict how to return to the previous location from the current viewpoint, as shown in Fig.\ref{fig:stmt}.
Given a trajectory $\{r_1, r_2, ..., r_t\}$, the model needs to predict $a_t^p$ in order to transit from $r_t$ back to $r_{t-1}$. We use $\texttt{prompt}_a$ to distinguish this task and introduce a new special token $\bm{x}^a_0$ for predicting $a^p_t$. The model input is:
\begin{equation}
    [r_1, r_2, ..., r_t;\text{\texttt{prompt}}_a,\bm{x}^a_0].
\end{equation}
We denote the output corresponding to $\bm{x}^a_0$ at the $l$-th LLM block as $\bm{x}^a_l \in \mathbb{R}^{1\times D_p}$. We then aggregate the visual features at step $t$ through an attention layer:
\begin{equation}
    \widetilde{\bm{x}}^a_l = \text{\texttt{cross\_attn}}(\bm{x}^a_l,\bm{I}_{t,1:36}).
\end{equation}
$\widetilde{\bm{x}}^a_l$ replaces $\bm{x}^a_l$ as the input for the following layers. To mitigate the impact on the primary model and enhance training stability, the aggregation operation only starts from the output of $L_s$-th LLM block. 
We replace the original word prediction layer with an attention mechanism to predict $a^p_t$:
\begin{equation}
    \bm{A}_t = \text{\texttt{softmax}}(\bm{x}_{L}^a \bm{W} \bm{I}^{\top}_{t,1:36}),
\end{equation}
where $\bm{W}\!\in\!\mathbb{R}^{D_p\times D_I\!}$ is a learnable projection matrix, $\bm{x}_{L}^a$ is the output of the LLM and $\bm{A}_t$ is the predicted distribution. We apply cross entropy loss over $\bm{A}_t$:
 \begin{equation}
     \mathcal{L}_a = \text{\texttt{cross\_entropy}}(a^p_t,\bm{A}_t).
 \end{equation}
During the training process, $\mathcal{L}_a$ is jointly optimized with the auto-regressive loss for instruction generation.

\subsection{Chain of Thought with Landmarks (CoTL)}
\label{sec:cotl}
Distinguished from image or video captioning, navigation instructions encompass more than just visual descriptions. 
An easily executable navigation instruction usually includes several \textbf{\textit{landmarks}} for directional guidance at crucial turning points. Besides, according to research in human cognitive psychology~\cite{lynch1964image}, it has been observed that humans, when providing path guidance, tend to first identify key navigation points within their cognitive maps before structuring their language. 
Therefore, the ability to determine landmarks is crucial for instruction generation. 
CoT~\cite{wei2022chain} has been validated as an effective means of guiding the reasoning process of LLMs.
Consequently, we introduce CoTL to direct the model to utilize critical landmarks in the navigation trajectory to generate instructions. 

\begin{figure*}[tb]
      \centering
      \includegraphics[width=1\textwidth]{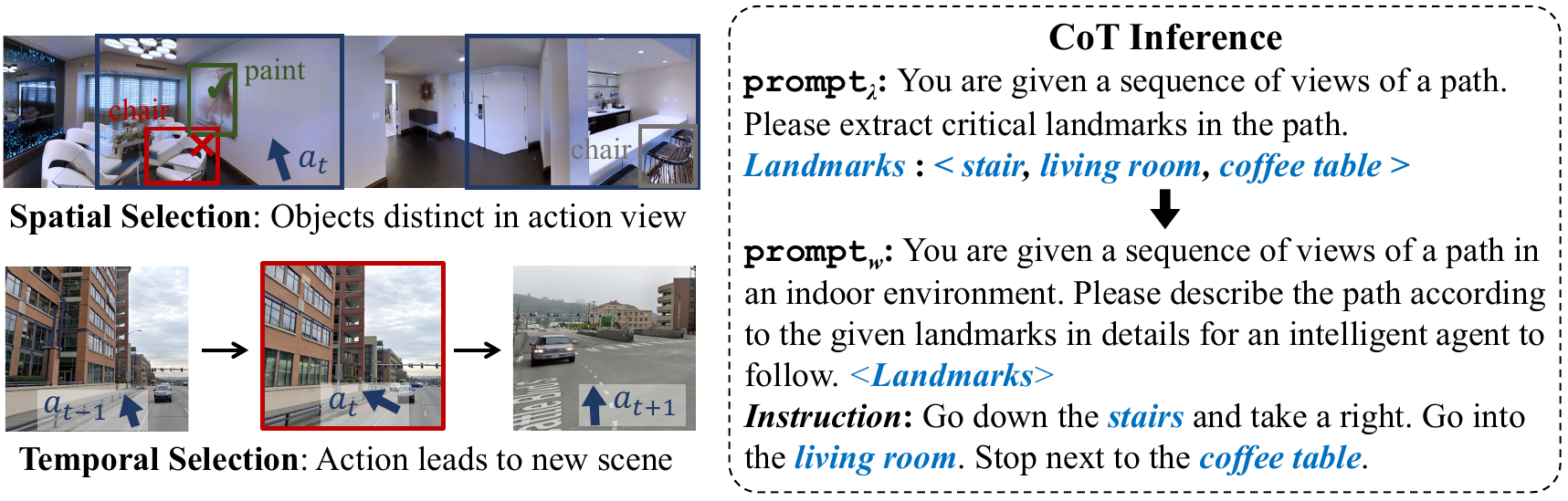}
      \caption{Details of Landmark Selection (left) and CoT Inference (right) in CoTL (\S\ref{sec:cotl}). In Spatial Selection, candidate views are partitioned in \textbf{\textcolor[RGB]{32,56,100}{blue boxes}}, and only objects that are distinct in action view are selected as landmarks (marked with a \textcolor[RGB]{56,87,35}{\textbf{green tick} \Checkmark}). In Temporal Selection, the action that leads to a new scene is treated as a significant viewpoint (marked in \textbf{\textcolor[RGB]{192,0,0}{red box}}).}
   \label{fig:cotl}
\end{figure*}

\noindent\textbf{Landmark Selection.} For the provided annotation pairs of instructions and paths in the training set, we initially extract nouns from the instructions as linguistic landmarks $\Lambda_x=\{\lambda^x_n\}_{n=1}^{N_x}$. Since valuable landmarks may not be fully specified in the annotated instructions, we supplement the landmark set by considering the visual characteristics of the path, as shown in \cref{fig:cotl}. We select visual landmarks from two perspectives, \ie, the temporal perspective and the spatial perspective. From the temporal perspective, we identify crucial viewpoints along the trajectory, where landmarks are more essential for guidance. Specifically, when the trajectory leads into a new scene, \eg, transitioning from a corridor to a room, the navigator often requires a landmark for guidance. We compute the feature difference of panoramic views along a trajectory to locate these viewpoints. For a given path, we construct a sequence comprising the mean-pooled features of panoramic views $\{\bm{I}^*_{t}\}^T_{t=1}$. We then compute the temporal importance score $\delta^\tau_t$ via cosine distance between $\bm{I}^*_{t}$ and $\bm{I}^*_{t+1}$:
\begin{equation}
    \delta^\tau_t = 1 - \frac{\bm{I}^*_{t} \cdot \bm{I}^*_{t+1}}{||\bm{I}^*_{t}||\cdot ||\bm{I}^*_{t+1}||},~~\bm{I}^*_{t} = \frac{1}{K}\sum_{k=1}^K \bm{I}_{t,k},
\end{equation}
where $\delta_t^\tau$ indicates the temporal importance of landmarks appearing at time step $t$. 
From the spatial perspective, we need to identify the most distinctive object to serve as a landmark. Distinctive objects are primarily the ones that appear in the action view and not in any other candidate views. At time step $t$, we first extract all objects appearing in $v_{t,a_t}$ as the candidate landmark set $\{\lambda^*_{t,n}\}_{n=1}^{N_t}$. Then, we assign distinctive scores according to the occurrence of landmarks in other candidate views. For example, the landmark  $\lambda^*_{t,n}$ that also appears in candidate views $\{c_1,c_2,c_3\}$ is assigned the spatial importance score $\delta^a_{t,n}$:
\begin{equation}
    \delta^a_{t,n} = 1 - d^a_{t,c_1} -  d^a_{t,c_2} -  d^a_{t,c_3},
\end{equation}
where $d^a_{t,c_i}$ is the cosine distance between view $a_t$ and view $c_i$. 
The final score for landmark $\lambda^*_{t,n}$ is:
\begin{equation}
    \delta_{t,n} = \delta^a_{t,n} \cdot \delta^\tau_t .
\end{equation}
We select landmarks with $\delta_{t,n}\ge\beta$ from all $\lambda^*_{t,n}$ in the trajectory to build the visual landmark set $\Lambda_v=\{\lambda^v_n\}_{n=1}^{N_v}$. Finally, the full landmark set of trajectory $\textit{R}$ can be constructed as:
\begin{equation}
    \Lambda=\Lambda_x \cup \Lambda_v.
\end{equation}

\noindent\textbf{CoT Training and Inference.} 
To enable the model to comprehensively identify landmarks, we utilize extracted landmarks $\Lambda$ to construct training data. For a trajectory $\textit{R}$, its corresponding data item consists of: 
\begin{equation}
 [\textit{R};\text{\texttt{prompt}}_{\lambda}, \Lambda],   
\end{equation}
where $\text{\texttt{prompt}}_{\lambda}$ is the prompt for landmark generation. During training, only the $\Lambda$ part is supervised. 

To equip the model with the ability to generate instructions according to given landmarks, the training data for instruction generation corresponding to a path $\textit{R}$ can be constructed as:  
\begin{equation}
[\textit{R};\text{\texttt{prompt}}_{w},\Lambda_x, \textit{X}],
\end{equation}
where only the $\textit{X}$ part is supervised during training. We establish a strong correspondence between landmarks and instructions in this phase by using only $\Lambda_x$ as the landmark input. This helps ensure the generation of diverse instructions by modifying landmarks.

Accordingly, the instruction generation process of the model (\cref{fig:cotl}) is divided into two stages. Firstly, given a trajectory $\textit{R}$, the model is guided by $\text{\texttt{prompt}}_{\lambda}$ to predict landmarks $M$. Then, using the generated $M$ and guided by $\text{\texttt{prompt}}_{w}$, the complete instruction is generated. 

There are two key advantages of this CoT paradigm. Firstly, it can highlight the landmarks within the path during training, enhancing the feasibility of instructions and reducing the risk of semantic errors in instruction generation. Secondly, by modifying the landmarks predicted in the first step, it allows for controlled alterations in the model's focus on landmarks in the trajectory. Further details of the prompts are discussed in the supplementary.

\subsection{Style-Mixed Training (SMT)}
\label{sec:smt}
In application, a model that can only generate step-by-step instructions is less practical.
When the instruction follower is familiar with the environment, fine-grained instructions lead to reduced communication efficiency. Additionally, due to the extensive amount of labor required for annotating navigation instructions, the data available is limited, especially for instructions with specified styles. This results in LLMs being susceptible to overfitting, makes it challenging to achieve accurate cross-modal mapping, and leads to suboptimal instruction generation performance when the model is trained with single-style instructions. 

To mitigate the issues above, we mix datasets with instructions in different linguistic styles for training. We devise descriptions that encapsulate diverse styles into prompts to enable the LLM to generate in different styles.
By employing SMT, not only is the quality of instruction generation enhanced, but we also enable a single LLM instance to adaptively generate different styles of instructions for the same path $\textit{R}$ by switching between different prompts.

\section{Experiments}
\label{sec:exp}
\subsection{Datasets and Evaluation Metrics}
\label{sec:dataset}
\noindent\textbf{Datasets.} 
We evaluate the instruction generation performance on three indoor navigation datasets~\cite{anderson2018vision,qi2020reverie,ku2020room} and one outdoor navigation dataset~\cite{huang2022assister}:
\begin{itemize}
\item R2R~\cite{anderson2018vision}: It has four splits with step-by-step instructions, \ie, \texttt{train} ($61$ scenes, $14,039$ instructions), \texttt{val} \texttt{seen} ($61$ scenes, $1,021$ instructions), \texttt{val} \texttt{unseen} ($11$ scenes, $2,349$ instructions), and \texttt{test} \texttt{unseen} ($18$ scenes, $4,173$ instructions).
 As \texttt{test} \texttt{unseen} is reserved for benchmarking instruction followers, we report the performance of instruction generation on \texttt{val} splits. 
 \item REVERIE~\cite{qi2020reverie}: It contains high-level descriptions of target destinations and objects. It has three open-access splits, \ie, \texttt{train} ($61$ scenes, $10,466$ instructions), \texttt{val} \texttt{seen} ($61$ scenes, $1,371$ instructions), and \texttt{val} \texttt{unseen} ($10$ scenes, $3,753$ instructions). We report the performance on two \texttt{val} splits.
 \item RxR~\cite{ku2020room}: It is a multilingual indoor navigation dataset with longer trajectories and more fine-grained aligned instructions. we specifically utilize the English instructions for comparison with previous methods. It has three publicly available splits, 
 and we report the performance on two \texttt{val} splits.
 \item UrbanWalk~\cite{huang2022assister}: It is an outdoor navigation dataset with $26,808$ image-instruction pairs simulated by CARLA~\cite{dosovitskiy2017carla}. We follow the setting in~\cite{Zeng_Wang_Wang_Yang_2023}.
\end{itemize}

The \texttt{val unseen} splits in R2R~\cite{anderson2018vision}, REVERIE~\cite{qi2020reverie}, and RxR~\cite{ku2020room} contain trajectories whose corresponding scenes are not included in \texttt{train} splits, and thus are good testbeds for generalizability~\cite{chen2022think, dou2022foam, Zeng_Wang_Wang_Yang_2023, zhang2023vln}. Consequently, we focus on those splits to better validate the generalizability of \textsc{C-Instructor}.

\noindent\textbf{Evaluation Metrics.} We evaluate the linguistic quality of generated instructions with widely-used automatic text similarity metrics, including BLEU~\cite{papineni2002bleu}, SPICE~\cite{anderson2016spice}, CIDEr~\cite{vedantam2015cider}, Meteor~\cite{banerjee2005meteor}, and Rouge~\cite{lin2004rouge}. 
For each navigation path, all corresponding ground-truth instructions are used as references.

\subsection{Implementation Details}
\label{sec:imp}
\noindent\textbf{Detailed Architecture.}
We use the multimodal LLaMA-Adapter~\cite{gao2023llama} with 32 layers and 7B parameters as the LLM. We adopt CLIP-ViT-L-14~\cite{radford2021learning} and 8 ViT~\cite{dosovitskiy2020image} blocks in the Trajectory Encoder. The score threshold $\beta$ for landmark selection in \S\ref{sec:cotl} is set to $0.25$, and $L_s$ in \S\ref{sec:stmt} is set to 30.

\noindent\textbf{Training.}
We only finetune the last 2 layers of LLM while fixing the other 30 layers. The CLIP~\cite{radford2021learning} visual encoder is also fixed. 
We first pre-train \textsc{C-Instructor} on PREVALENT~\cite{hao2020towards} for 240K iterations with a batch size of 16, and then fine-tune \textsc{C-Instructor} on multiple datasets jointly for 120K iterations with batch size 4. We use the AdamW~\cite{loshchilov2017decoupled} optimizer with base learning rate $1.0\times 10^{-4}$. Four NVIDIA A100 80GB GPUs are used for training.

\noindent\textbf{Inference.}
We set the generation temperature to $1.0$ for RxR~\cite{ku2020room}, and $0.1$ for all other datasets. 
All other hyperparameters remain the same as~\cite{gao2023llama}.


\subsection{Comparison to State-of-the-Art Methods}
\label{sec:compare}

We compare \textsc{C-Instructor} with four existing instruction generation models. 
For a fair comparison, we report the performance of \textsc{C-Instructor} without SMT in addition to the performance of the full model.
We employ the Penn Treebank tokenizer~\cite{taylor2003penn} to compute the linguistic metrics.

\begin{table*}[tb]
		\caption{Comparison to state-of-the-art methods (\S\ref{sec:compare}) on R2R~\cite{anderson2018vision}.}
	\centering
			\resizebox{1\textwidth}{!}{
			\setlength\tabcolsep{2pt}
			\renewcommand\arraystretch{1.25}
	\begin{tabular}{|rl||cccccc|cccccc|}
	\hline \thickhline
	~ & & \multicolumn{6}{c|}{R2R \texttt{val} \texttt{seen}} & \multicolumn{6}{c|}{R2R \texttt{val} \texttt{unseen}}\\
	\cline{3-14}\cline{3-14}\cline{3-14}
	\multicolumn{2}{|c||}{\multirow{-2}{*}{Methods}}
	&\texttt{SPICE}\!~$\uparrow$ &\texttt{BLEU-1}\!~$\uparrow$ &\texttt{BLEU-4}\!~$\uparrow$ &\texttt{CIDEr}\!~$\uparrow$ &\texttt{Meteor}\!~$\uparrow$ &\texttt{Rouge}\!~$\uparrow$ &\texttt{SPICE}\!~$\uparrow$  &\texttt{BLEU-1}\!~$\uparrow$ &\texttt{BLEU-4}\!~$\uparrow$ &\texttt{CIDEr}\!~$\uparrow$ &\texttt{Meteor}\!~$\uparrow$ &\texttt{Rouge}\!~$\uparrow$  \\
	\hline
	\hline
	BT-speaker$_{\!}$~\cite{Fried_Hu_Cirik_Rohrbach_Andreas_Morency_Berg-Kirkpatrick_Saenko_Klein_Darrell_2018}&\!\!\pub{NeurIPS2018}   & 0.173 & 0.670 & 0.236 & 0.373 & 0.209 &0.443 & 0.113 & 0.600 & 0.149 & 0.113 & 0.167 & 0.376	\\
    EDrop-speaker$_{\!}$~\cite{tan2019learning}&\!\!\pub{NAACL2019}   & 0.168 & 0.660 & 0.228 & 0.362 & 0.208 & 0.447 & 0.117 & 0.590 & 0.157  & 0.160	& 0.174 & 0.389 	 \\
	 CCC-speaker$_{\!}$~\cite{wang2022counterfactual}&\!\!\pub{CVPR2022} & 0.194 & 0.698 & 0.265 & 0.449 & 0.218 & 0.467 & 0.108 & 0.591 & 0.139 & 0.120 & 0.164 & 0.375   \\
  Lana$_{\!}$~\cite{wang2023lana}&\!\!\pub{CVPR2023} & 0.170 & 0.657 & 0.215 & 0.265 & 0.205 & 0.433 & 0.174 & 0.667 & 0.236 & 0.295 & 0.213 & 0.448 \\
  \hline
    \textbf{\textsc{C-Instructor}}&\textit{w/o} SMT& 0.230 & \textbf{0.732} & 0.270 & 0.511 & 0.237 & 0.475 & \textbf{0.217} & \textbf{0.715} & 0.263 & \textbf{0.453} & 0.234 & 0.470 \\
    \textbf{\textsc{C-Instructor}}&(Ours)& \textbf{0.233} & 0.726 & \textbf{0.276} & \textbf{0.529} & \textbf{0.247} & \textbf{0.480} & 0.212 & 0.713 & \textbf{0.266} & 0.447 & \textbf{0.239} & \textbf{0.473} \\
\hline
	\end{tabular}
	}
		\label{table:r2r_g}
\end{table*}
\begin{table*}[tb]
		\caption{Comparison to state-of-the-art methods (\S\ref{sec:compare}) on REVERIE~\cite{qi2020reverie}.}
	\centering
			\resizebox{1\textwidth}{!}{
			\setlength\tabcolsep{2pt}
			\renewcommand\arraystretch{1.25}
	\begin{tabular}{|rl||cccccc|cccccc|}
	\hline \thickhline
	~ & & \multicolumn{6}{c|}{REVERIE \texttt{val} \texttt{seen}} & \multicolumn{6}{c|}{REVERIE \texttt{val} \texttt{unseen}}\\
	\cline{3-14}\cline{3-14}\cline{3-14}
	\multicolumn{2}{|c||}{\multirow{-2}{*}{Methods}}
	&\texttt{SPICE}\!~$\uparrow$ &\texttt{BLEU-1}\!~$\uparrow$ &\texttt{BLEU-4}\!~$\uparrow$ &\texttt{CIDEr}\!~$\uparrow$ &\texttt{Meteor}\!~$\uparrow$ &\texttt{Rouge}\!~$\uparrow$ &\texttt{SPICE}\!~$\uparrow$  &\texttt{BLEU-1}\!~$\uparrow$ &\texttt{BLEU-4}\!~$\uparrow$ &\texttt{CIDEr}\!~$\uparrow$ &\texttt{Meteor}\!~$\uparrow$ &\texttt{Rouge}\!~$\uparrow$  \\
	\hline
	\hline
	BT-speaker$_{\!}$~\cite{Fried_Hu_Cirik_Rohrbach_Andreas_Morency_Berg-Kirkpatrick_Saenko_Klein_Darrell_2018}&\!\!\pub{NeurIPS2018}   & 0.121 & 0.693 & 0.347 & 0.269 & 0.223 &0.602 & 0.103 & 0.664 & 0.302 & 0.190& 0.200 & 0.569\\
    EDrop-speaker$_{\!}$~\cite{tan2019learning}&\!\!\pub{NAACL2019}   & 0.138 & 0.641 & 0.360 & 0.523 & 0.277 & 0.597 & 0.114 & 0.648 & 0.319  & 0.333	& 0.233 & 0.546	 \\
 CCC-speaker$_{\!}$~\cite{wang2022counterfactual}&\!\!\pub{CVPR2022} & 0.144 & 0.727 & 0.408 & 0.502 & 0.272 & 0.589 & 0.115 & 0.681 & 0.357 & 0.334 & 0.232 & 0.548   \\
  Lana$_{\!}$~\cite{wang2023lana}&\!\!\pub{CVPR2023} & 0.137 & 0.707 & 0.404 & 0.627 & 0.282 & 0.619 & 0.107 & 0.696 & 0.345 & 0.327 & 0.239 & 0.582 \\
  \hline
  \textbf{\textsc{C-Instructor}} &\textit{w/o} SMT& \textbf{0.184} & \textbf{0.785} & \textbf{0.480} & \textbf{0.844} & \textbf{0.319} & \textbf{0.649} & 0.139 & 0.739 & 0.369 & 0.464 & 0.259 & 0.577 \\
  \textbf{\textsc{C-Instructor}} & (Ours)& 0.182 & 0.775 & 0.459 & 0.805 & 0.311 & 0.647 & \textbf{0.141} & \textbf{0.754} & \textbf{0.419} & \textbf{0.545} & \textbf{0.267} & \textbf{0.591} \\
\hline
	\end{tabular}
	}
		\label{table:reverie_g}
\end{table*}
\begin{table*}[tb]
		\caption{Comparison to state-of-the-art methods (\S\ref{sec:compare}) on RxR~\cite{ku2020room}.}
	\centering
			\resizebox{1\textwidth}{!}{
			\setlength\tabcolsep{2pt}
			\renewcommand\arraystretch{1.2}
	\begin{tabular}{|rl||ccccc|ccccc|}
	\hline \thickhline
	~ & & \multicolumn{5}{c|}{RxR \texttt{val} \texttt{seen}} & \multicolumn{5}{c|}{RxR \texttt{val} \texttt{unseen}}\\
	\cline{3-12}\cline{3-12}\cline{3-12}
	\multicolumn{2}{|c||}{\multirow{-2}{*}{Methods}}
	&\texttt{BLEU-1}\!~$\uparrow$ &\texttt{BLEU-4}\!~$\uparrow$ &\texttt{CIDEr}\!~$\uparrow$ &\texttt{Meteor}\!~$\uparrow$ &\texttt{Rouge}\!~$\uparrow$ &\texttt{BLEU-1}\!~$\uparrow$ &\texttt{BLEU-4}\!~$\uparrow$ &\texttt{CIDEr}\!~$\uparrow$ &\texttt{Meteor}\!~$\uparrow$ &\texttt{Rouge}\!~$\uparrow$  \\
	\hline
	\hline
	BT-speaker$_{\!}$~\cite{Fried_Hu_Cirik_Rohrbach_Andreas_Morency_Berg-Kirkpatrick_Saenko_Klein_Darrell_2018}&\!\!\pub{NeurIPS2018}   & 0.514 & 0.188 & 0.026 & 0.204 & 0.365 &0.566 & 0.211 & 0.024 & 0.208 & 0.372 	\\
    EDrop-speaker$_{\!}$~\cite{tan2019learning}&\!\!\pub{NAACL2019}   & 0.595 & 0.197 & 0.047 & 0.213 & 0.378 & 0.568 & 0.184 & 0.038 & 0.205  & 0.370		 \\
	 CCC-speaker$_{\!}$~\cite{wang2022counterfactual}&\!\!\pub{CVPR2022} & 0.526 & 0.194 & 0.024 & 0.185 & 0.355 & 0.518 & 0.187 & 0.026 & 0.184 & 0.353    \\
  Lana$_{\!}$~\cite{wang2023lana}&\!\!\pub{CVPR2023} & 0.342 & 0.123 & 0.040 & 0.128 & 0.275 & 0.319 & 0.115 & 0.043 & 0.124 & 0.273  \\
  \hline
  \textbf{\textsc{C-Instructor}} &\textit{w/o} SMT& 0.683 & 0.233 & 0.081 & \textbf{0.243} & 0.381 & 0.667 & 0.224 & \textbf{0.080} & 0.236 & 0.379 \\
  \textbf{\textsc{C-Instructor}} &(Ours)& \textbf{0.685} & \textbf{0.234} & \textbf{0.082} & 0.238 & \textbf{0.382} & \textbf{0.678} & \textbf{0.233} & 0.077 & \textbf{0.239} & \textbf{0.382} \\
\hline
	\end{tabular}
	}
		\label{table:rxr_g}
\end{table*}
\begin{table}[tb]
\caption{Comparison to state-of-the-art methods (\S\ref{sec:compare}) on UrbanWalk~\cite{huang2022assister}.}
\centering
\resizebox{0.85\linewidth}{!}{
			\setlength\tabcolsep{5pt}
   \renewcommand\arraystretch{1.2}
\begin{tabular}{|rl||ccccc|} 
\hline\thickhline
\multicolumn{2}{|c||}{\multirow{2}{*}{Methods}} & \multicolumn{5}{c|}{UrbanWalk}                                                                                                                               \\ 
\cline{3-7}
\multicolumn{2}{|c||}{}                         & \multicolumn{1}{l}{\texttt{SPICE}~$\uparrow$} & \multicolumn{1}{l}{\texttt{BLEU-1}~$\uparrow$} & \multicolumn{1}{l}{\texttt{BLEU-4}~$\uparrow$} & \multicolumn{1}{l}{\texttt{Meteor}~$\uparrow$} & \multicolumn{1}{l|}{\texttt{Rouge}~$\uparrow$}  \\ 
\hline\hline
BT-speaker$_{\!}$~\cite{Fried_Hu_Cirik_Rohrbach_Andreas_Morency_Berg-Kirkpatrick_Saenko_Klein_Darrell_2018}&\pub{NeurIPS2018}                                  & 0.524                              & 0.649 & 0.408                                 & 0.350                                  & 0.620                                   \\
EDrop-speaker$_{\!}$~\cite{tan2019learning}&\pub{NAACL2019}                                 & 0.531                              & 0.689 & 0.435                                 & 0.358                                 & 0.634                                  \\
ASSISTER$_{\!}$~\cite{huang2022assister}   & \pub{ECCV2022}                                 & 0.451                              & 0.576 & 0.164                                 & 0.319                                 & 0.557                                  \\
Kefa-speaker$_{\!}$~\cite{Zeng_Wang_Wang_Yang_2023} & \pub{Arxiv2023}                                 & 0.566                              & 0.711 & 0.450                                 & 0.378                                 & 0.655                                  \\
\hline
\textbf{\textsc{C-Instructor}}       & (Ours)                             & \textbf{0.645}                              & \textbf{0.771} & \textbf{0.534}                                 & \textbf{0.461}                                 & \textbf{0.781}                                  \\
\hline
\end{tabular}}
		\label{table:ubw_g}
\end{table}

\noindent\textbf{R2R~\cite{anderson2018vision}.} The results on R2R are summarized in \cref{table:r2r_g}. \textsc{C-Instructor} outperforms previous methods under all metrics on both \texttt{val} splits. 
In terms of SPICE, \textsc{C-Instructor} demonstrates a superiority of $3.9\%$ in absolute terms and $20.1\%$ in relative terms on \texttt{val} \texttt{seen} as well as $3.8\%$ in absolute terms and $21.8\%$ in relative terms on \texttt{val} \texttt{unseen} compared to the previous best. 
This verifies that \textsc{C-Instructor} exhibits good performance in generating fine-grained directives.

\noindent\textbf{REVERIE~\cite{qi2020reverie}.} 
As depicted in \cref{table:reverie_g}, \textsc{C-Instructor} also attains state-of-the-art performance in generating high-level trajectory descriptions. It exhibits a relative improvement of $26.4\%$ on \texttt{val} \texttt{seen} and $22.6\%$ on \texttt{val} \texttt{unseen} in terms of SPICE, which is more pronounced compared to R2R~\cite{anderson2018vision}. 

\noindent\textbf{RxR~\cite{ku2020room}.} 
As shown in \cref{table:rxr_g}, \textsc{C-Instructor} significantly outperforms existing instruction generation algorithms in all metrics. 
This suggests that \textsc{C-Instructor} possesses the capability to manage visual contexts of extended trajectory and generate more intricate instructions.

\noindent\textbf{UrbanWalk~\cite{huang2022assister}.} As shown in~\cref{table:ubw_g}, \textsc{C-Instructor} also achieves the best performance under all metrics on outdoor scenes. This indicates that our \textsc{C-Instructor} possesses strong generalization capability and universality.


\subsection{Diagnostic Experiment}
\label{sec:ablation}

\begin{table*}[tb]
		\caption{Ablation study (\S\ref{sec:ablation}) on REVERIE~\cite{qi2020reverie} \texttt{val} \texttt{unseen} and R2R~\cite{anderson2018vision} \texttt{val} \texttt{unseen}.}
	\centering
			\resizebox{1\textwidth}{!}{
			\setlength\tabcolsep{2pt}
			\renewcommand\arraystretch{1.2}
	\begin{tabular}{|rl||ccccc|ccccc|}
	\hline \thickhline
	~ & & \multicolumn{5}{c|}{REVERIE \texttt{val} \texttt{unseen}} & \multicolumn{5}{c|}{R2R \texttt{val} \texttt{unseen}}\\
	\cline{3-12}\cline{3-12}\cline{3-12}
	\multicolumn{2}{|c||}{\multirow{-2}{*}{Methods}}
	&\texttt{BLEU-1}\!~$\uparrow$ &\texttt{BLEU-4}\!~$\uparrow$ &\texttt{CIDEr}\!~$\uparrow$ &\texttt{Meteor}\!~$\uparrow$ &\texttt{Rouge}\!~$\uparrow$ &\texttt{BLEU-1}\!~$\uparrow$ &\texttt{BLEU-4}\!~$\uparrow$ &\texttt{CIDEr}\!~$\uparrow$ &\texttt{Meteor}\!~$\uparrow$ &\texttt{Rouge}\!~$\uparrow$  \\
	\hline
	\hline
    \multicolumn{2}{|l||}{Vanilla LLM} & 0.399 & 0.131 & 0.432 & 0.156 & 0.400 &  0.307 & 0.059 & 0.292 & 0.139 & 0.303	\\
    Baseline&    & 0.648 & 0.308 & 0.347 & 0.248 & 0.547 &  0.676 & 0.232 & 0.356 & 0.225 & 0.449	\\
    Baseline&+ SMT   & 0.679 & 0.344   & 0.397 & 0.254 & 0.562 & 0.685 & 0.254  & 0.407	& 0.233 & 0.466 	 \\
  Baseline&+ SMT + STMT & 0.737 & 0.402 & 0.490 & 0.258 & 0.590 & 0.689 & 0.262 & 0.445 & 0.228 & \textbf{0.479} \\
    Baseline&+ SMT + STMT + CoTL & \textbf{0.754} & \textbf{0.419} & \textbf{0.545} & \textbf{0.267} & \textbf{0.591} & \textbf{0.713} & \textbf{0.266} & \textbf{0.447} & \textbf{0.239} & 0.473 \\
  \hline
 
	\end{tabular}
	}
		\label{table:ablation}
\end{table*}

To thoroughly study the effectiveness of \textsc{C-Instructor}, we compare the full model with several ablative designs. We test the ablative models on REVERIE~\cite{qi2020reverie} and R2R~\cite{anderson2018vision} \texttt{val} \texttt{unseen}. The results are summarized in~\cref{table:ablation}. 

\noindent\textbf{Vanilla LLM.} We assess the performance of vanilla LLM by captioning views along the trajectory using BLIP~\cite{li2022blip} and feeding those captions with devised prompts into pre-trained LLaMA~\cite{gao2023llama} to generate navigation instructions. The performance of this vanilla method fine-tuned on REVERIE~\cite{qi2020reverie} and R2R~\cite{anderson2018vision} respectively (\#1) remains largely inferior to the baseline in \S\ref{sec:framework} (\#2), which still significantly lags behind our full method (\#5). This underscores the inherent information loss through captioning as well as the effectiveness of our design.

\noindent\textbf{SMT.} To train a model with instructions from diverse domains yields performance benefits. In comparison to \#2, the model trained using SMT (\#3) exhibits an improvement in SPICE on the REVERIE \texttt{val} \texttt{unseen} from $0.127$ to $0.129$. It concurrently achieves a performance improvement on the R2R \texttt{val} \texttt{unseen}. This suggests that enhancing linguistic diversity will foster the quality of instructions generated by \textsc{C-Instructor}. 

\noindent\textbf{STMT.} The model trained with STMT (\#4) demonstrates a notable impact on generating highly abstract instructions. It lifts BLEU-4 from $0.344$ to $0.402$ and CIDEr from $0.397$ to $0.490$ on the REVERIE \texttt{val} \texttt{unseen}. This highlights the significance of understanding the environment layout.

\noindent\textbf{CoTL.} Compared to \#4, the model with CoTL (\#5) significantly improves the semantic consistency with the ground truth instruction. The improvement on REVERIE is more significant: SPICE increases from $0.129$ to $0.141$. This suggests that incorporating CoTL enhances the alignment between generated instructions and the visual environment, especially for high-level instructions.

\subsection{Instruction Quality Analysis} 
\label{sec:navigation}

\begin{table*}[tb]
\caption{Instruction quality analysis based on performance of navigation models (\S\ref{sec:navigation}).}
\centering
\resizebox{1\linewidth}{!}
{
\captionsetup{font=large, position=bottom}
\subfloat[Performance of HAMT~\cite{chen2021history} using different instruction generator for data augmentation on REVERIE~\cite{qi2020reverie} \texttt{val} \texttt{unseen} (\S\ref{sec:navigation}). Training with instructions generated by \textsc{C-Instructor} yields the most significant improvement. \label{table:nav_aug}]{
\setlength\tabcolsep{4pt}
   \renewcommand\arraystretch{1.35}

\begin{tabular}{|l||cccc|}
\hline \thickhline 
\multicolumn{1}{|c||}{\multirow{2}{*}{Data Source}} & \multicolumn{4}{c|}{REVERIE \texttt{val} \texttt{unseen}}  \\ 
\cline{2-5}
\multicolumn{1}{|c||}{}       & \texttt{SR}\!~$\uparrow$    & \texttt{SPL}\!~$\uparrow$   & \texttt{RGS}\!~$\uparrow$   & \texttt{RGSPL}\!~$\uparrow$    \\ 
\hline\hline
~~Original~\cite{qi2020reverie}&  32.95 & 30.20  & 18.92 & 17.28    \\
+BT-speaker~\cite{Fried_Hu_Cirik_Rohrbach_Andreas_Morency_Berg-Kirkpatrick_Saenko_Klein_Darrell_2018} & 31.84 & 28.37 & 17.35 & 15.14    \\
+EDrop-speaker~\cite{tan2019learning}& 30.45 & 27.18 & 18.60  & 16.24    \\
+CCC-speaker~\cite{wang2022counterfactual}& 29.65 & 26.20 & 16.33 & 14.58 \\
+Lana~\cite{wang2023lana}& 33.05 & 29.76 & 19.14 & 17.20    \\ 
\hline
+\textbf{\textsc{C-Instructor}} (Ours) & \textbf{34.25} & \textbf{31.25} & \textbf{19.99} & \textbf{18.08}    \\
\hline
\end{tabular}
}
\hspace{3mm}
\subfloat[Performance of HAMT~\cite{chen2021history} and DUET~\cite{chen2022think} in following instructions generated on REVERIE~\cite{qi2020reverie} \texttt{val} \texttt{unseen} (\S\ref{sec:navigation}). SR and SPL are provided as metrics to evaluate the path-guiding proficiency of different instruction generation models.\label{tab:mnav}]{
\setlength\tabcolsep{5pt}
   \renewcommand\arraystretch{1.2}
\begin{tabular}{|l||cc|cc|} 
\hline \thickhline
\multicolumn{1}{|c||}{\multirow{3}{*}{Instruction Generator}} & \multicolumn{4}{c|}{Follower}                                                                             \\ 
\cline{2-5}
\multicolumn{1}{|c||}{}                        & \multicolumn{2}{c|}{HAMT~\cite{chen2021history}}                         & \multicolumn{2}{c|}{DUET~\cite{chen2022think}}                             \\ 
\cline{2-5}
\multicolumn{1}{|c||}{}                        & \multicolumn{1}{c}{\texttt{SR}\!~$\uparrow$} & \multicolumn{1}{c|}{ \texttt{SPL}\!~$\uparrow$} & \multicolumn{1}{c}{\texttt{SR}\!~$\uparrow$}    & \multicolumn{1}{l|}{\texttt{SPL}\!~$\uparrow$}  \\ 
\hline\hline
\rowcolor[rgb]{0.853,0.853,0.853} Human annotation~\cite{qi2020reverie}& 32.95                  & 30.20                     & 46.98                     & 33.73                     \\ 
BT-speaker$_{\!}$~\cite{Fried_Hu_Cirik_Rohrbach_Andreas_Morency_Berg-Kirkpatrick_Saenko_Klein_Darrell_2018} & 24.85                  & 21.74                    & 30.47                     & 21.46                     \\
EDrop-speaker$_{\!}$~\cite{tan2019learning}& 26.19                  & 23.55                    & 27.89                     & 17.00                     \\
 CCC-speaker$_{\!}$~\cite{wang2022counterfactual}& 23.29                  & 20.69                    & 29.74                     & 19.55                     \\
  Lana$_{\!}$~\cite{wang2023lana} & 26.84 & 24.38 & 31.39 & 20.44 \\
  \hline
  \textbf{\textsc{C-Instructor}} (Ours) & \textbf{31.35}                  & \textbf{29.27}                    & \textbf{43.34}                     & \textbf{30.13}                     \\
\hline
\end{tabular}
}
}
\end{table*}

Evaluating the quality of instructions solely based on text similarity metrics is insufficient as those metrics do not thoroughly assess the semantic alignment between instructions and trajectories. Thus, we further analyze the semantic quality of instructions generated by \textsc{C-Instructor} from three aspects through the following experiments:

\noindent\textbf{Path Guiding Proficiency.} The success rate (SR) of navigators with instructions from different instruction generators can be used as an index for the quality of instructions. We regenerate instructions for the paths in REVERIE~\cite{qi2020reverie} \texttt{val} \texttt{unseen} and employ two navigators (HAMT~\cite{chen2021history} and DUET~\cite{chen2022think}) to assess SR and SPL (SR weighted by Path Length) when guided by regenerated instructions. 
As depicted in \cref{tab:mnav}, SR and SPL of instructions provided by \textsc{C-Instructor} significantly exceeds that of those generated by prior models and remarkably aligns with the navigation accuracy of human-annotated instructions.

\noindent\textbf{Data Augmentation.} The enhancement of navigation accuracy of instruction followers via data augmentation can also serve as an indicator for the improved quality of instruction generation. Hence, we leverage $17,533$ instructions generated by various instruction generation models on randomly sampled paths along with the original \texttt{train} split of REVERIE~\cite{qi2020reverie} to train HAMT~\cite{chen2021history}. As shown in~\cref{table:nav_aug}, the model utilizing data generated by \textsc{C-Instructor} exhibits an increase in the accuracy of navigation including SR, SPL, RGS (Remote Grounding Success rate), and RGSPL (RGS weighted by Path Length). 
RGS and RGSPL measure the success rate of the agent's finding the target object indicated in the given instruction and are used as navigator performance metrics on REVERIE~\cite{qi2020reverie}. 
In contrast, employing other models for data augmentation results in an unintended performance drop for the navigator. This indicates that \textsc{C-Instructor}, when utilized as a means of data augmentation, exhibits superior efficacy in generating instructions with high-level abstraction.

\noindent\textbf{User Study.} 
To provide a more comprehensive evaluation of the semantic quality of generated instructions, we conduct a series of human evaluations. Specifically, 15 college students are individually tasked with scoring from 0 to 5 according to the semantic alignment between the given instructions and the corresponding trajectories. The instructions provided are generated by \textsc{C-Instructor}, Lana~\cite{wang2023lana}, CCC~\cite{wang2022counterfactual}, BT-Speaker~\cite{Fried_Hu_Cirik_Rohrbach_Andreas_Morency_Berg-Kirkpatrick_Saenko_Klein_Darrell_2018}, and EnvDrop-Speaker~\cite{tan2019learning} from a total of 100 paths. The paths are sampled from the \texttt{val} \texttt{unseen} split of REVERIE~\cite{qi2020reverie}. \textsc{C-Instructor} garners a higher average score, \ie, $3.50$, \textit{vs} Lana $2.26$, CCC $2.14$, BT-Speaker $2.10$ and EnvDrop-Speaker $2.10$. 
This result further validates that the instructions generated by \textsc{C-Instructor} are well aligned with the corresponding navigation paths. 

\begin{figure*}[tb]
      \centering
      \includegraphics[width=\textwidth]{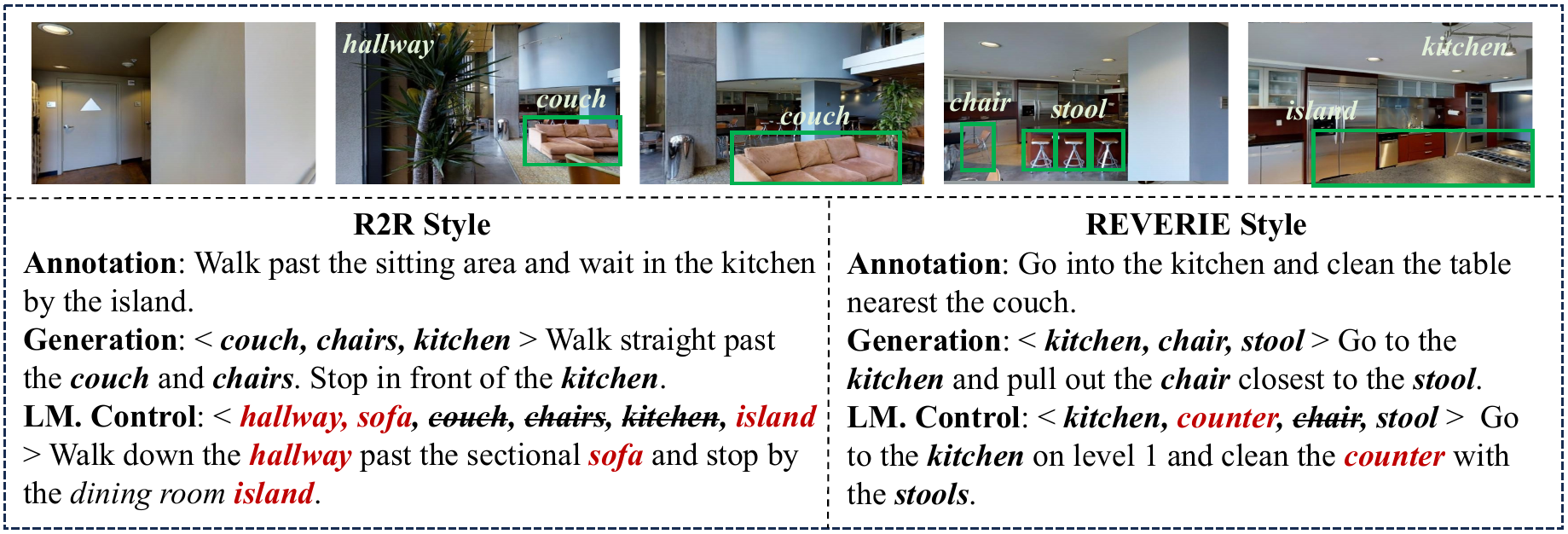}
      \caption{Visualizations of navigation trajectory and instruction generation results on R2R~\cite{anderson2018vision} and REVERIE~\cite{qi2020reverie} (\S\ref{sec:qual}).}
   \label{fig:vis}
\end{figure*}

\begin{figure}[tb]
      \centering
      \includegraphics[width=\linewidth]{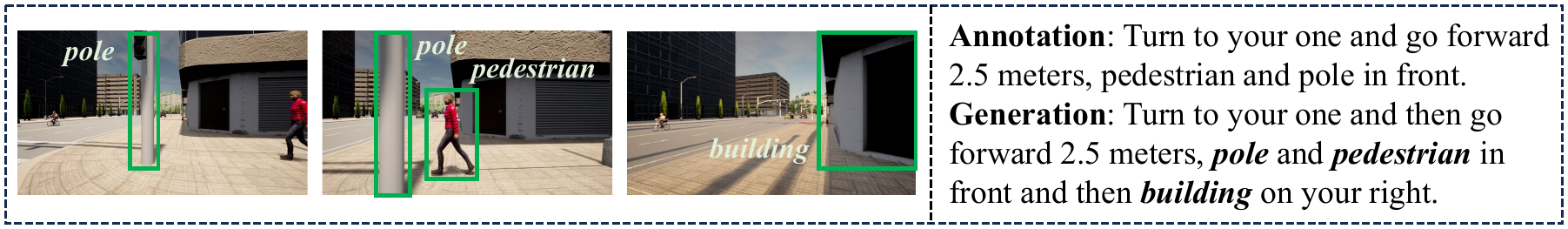}
      \caption{Visualizations of path and generated instruction on UrbanWalk~\cite{huang2022assister} (\S\ref{sec:qual}).}
   \label{fig:vis_ubw}
\end{figure}

\subsection{Qualitative Results}
\label{sec:qual}
We visualize an example of indoor navigation trajectory and corresponding instruction generation results in \cref{fig:vis}. As seen, \textsc{C-Instructor} can identify critical landmarks in the path and generate high-quality instructions accordingly in specified styles. Moreover, we can control the focus of \textsc{C-Instructor} by modifying landmarks. \cref{fig:vis_ubw} displays a result on UrbanWalk~\cite{huang2022assister}. We can observe that \textsc{C-Instructor} can also provide practical instructions in outdoor scenes.

\section{Conclusion and Discussion}
\label{sec:conclusion}

In this work, we propose \textsc{C-Instructor}, which generates style-controllable and content-controllable instructions with high linguistic quality. It uses an adapter-based structure to leverage the language capability of LLMs and distinct style prompts in SMT to achieve style control. To enhance the executability of generated instructions, we adopt CoTL to help identify crucial landmarks and provide content controllability. We also devise STMT to enhance the model's understanding of the environment's spatial topology. 
The instructions generated by \textsc{C-Instructor} not only achieve high scores in text metrics but also demonstrate strong competence in guiding navigators, further validating the strong correspondence between generated instructions and given trajectories. 
We expect that \textsc{C-Instructor} can greatly enhance agent-human communication and significantly contribute to the development of versatile embodied agents. 



%
%
\bibliographystyle{splncs04}
\bibliography{main}

\clearpage
\maketitlesupplementary
\appendix

This document provides more details, extra experimental results, and further discussion of \textsc{C-Instructor}. The document is organized as follows:
\begin{itemize}
    \item \S\ref{sec:prompts} provides detailed prompts for several datasets.
    \item \S\ref{sec:ablation_extra} presents extra ablation results on different selection strategies and values of $\beta$ in landmark selection.
    \item \S\ref{sec:stmt_analysis} further analyzes the effect of STMT through the training process.
    \item \S\ref{sec:extra_vis} shows more qualitative results of instruction generation and analyzes some failure cases.
    \item \S\ref{sec:discussion} discusses the social impact and limitations of our work, and suggests potential future work.
\end{itemize}

\section{Detailed Prompts}
\label{sec:prompts}

In this section, we provide detailed prompts for different navigation datasets. Note that all the given prompts are then formatted by prompt templates in \cite{gao2023llama}.

\begin{itemize}
    \item $\texttt{prompt}_{\lambda}$ for R2R~\cite{anderson2018vision}: You are given a sequence of views of a path. Please extract critical landmarks in the path.
    \item $\texttt{prompt}_{w}$ for R2R~\cite{anderson2018vision}: You are given a sequence of views of a path in an indoor environment. Please describe the path according to the given landmarks in detail for an intelligent agent to follow. Landmarks: $<$\textit{landmarks}$>$.
    \item $\texttt{prompt}_{\lambda}$ for REVERIE~\cite{qi2020reverie}: You are given a sequence of views of a path in an indoor environment. Please extract several critical landmarks in the path for generating a brief high-level target-oriented instruction.
    \item $\texttt{prompt}_{w}$ for REVERIE~\cite{qi2020reverie}: You are given a sequence of views of a path in an indoor environment and critical landmarks for a brief high-level target-oriented instruction. Please generate the indicated high-level target-oriented instruction briefly for an intelligent agent to follow. Landmarks: $<$\textit{landmarks}$>$.
    \item $\texttt{prompt}_{\lambda}$ for RxR~\cite{ku2020room}: You are given a sequence of views of a path in an indoor environment. Please extract critical landmarks describing the starting position and the path.
    \item $\texttt{prompt}_{w}$ for RxR~\cite{ku2020room}: You are given a sequence of views of a path in an indoor environment. Please describe the starting position and the path according to the given landmarks in detail for an intelligent agent to follow. Landmarks: $<$\textit{landmarks}$>$.
    \item $\texttt{prompt}_{a}$: You are an intelligent embodied agent that navigates in an indoor environment. Your task is to move among the static viewpoints (positions) of a pre-defined graph of the environment. You are given several candidate views. You are also given a sequence of panoramic views showing previous steps you have taken and the previous viewpoint you should return to. Now you should make an action by selecting a candidate view to return to the previous viewpoint. Candidate Views: $<$\textit{viewpoints}$>$
\end{itemize}










\section{Extra Ablations on Landmark Selection}
\label{sec:ablation_extra}

\subsection{Selection Strategies}
\label{sec:ablation_lm}

\begin{table*}[tb]
		\caption{Ablations on landmark selection strategies (\S\ref{sec:ablation_lm}) on REVERIE~\cite{qi2020reverie} \texttt{val} \texttt{unseen} and R2R~\cite{anderson2018vision} \texttt{val} \texttt{unseen}.}
	\centering
			\resizebox{1\textwidth}{!}{
			\setlength\tabcolsep{2pt}
			\renewcommand\arraystretch{1.25}
	\begin{tabular}{|rl||cccccc|cccccc|}
	\hline \thickhline
	~ & & \multicolumn{6}{c|}{REVERIE \texttt{val} \texttt{unseen}} & \multicolumn{6}{c|}{R2R \texttt{val} \texttt{unseen}}\\
	\cline{3-14}\cline{3-14}\cline{3-14}
	\multicolumn{2}{|c||}{\multirow{-2}{*}{Methods}}
	&\texttt{SPICE}\!~$\uparrow$ &\texttt{BLEU-1}\!~$\uparrow$ &\texttt{BLEU-4}\!~$\uparrow$ &\texttt{CIDEr}\!~$\uparrow$ &\texttt{Meteor}\!~$\uparrow$ &\texttt{Rouge}\!~$\uparrow$ &\texttt{SPICE}\!~$\uparrow$  &\texttt{BLEU-1}\!~$\uparrow$ &\texttt{BLEU-4}\!~$\uparrow$ &\texttt{CIDEr}\!~$\uparrow$ &\texttt{Meteor}\!~$\uparrow$ &\texttt{Rouge}\!~$\uparrow$  \\
	\hline
	\hline
  Baseline& & 0.129 & 0.737 & 0.402 & 0.490 & 0.258 & 0.590 & 0.194 & 0.689 & 0.262 & 0.445 & 0.228 & \textbf{0.479} \\
  Baseline& + $\Lambda_x$ & 0.143 & 0.732 & 0.380 & 0.482 & 0.263 & 0.580 & 0.199 & 0.687 & 0.252 & 0.416 & 0.230 & 0.466 \\
  Baseline& + $\Lambda_x \cup \Lambda_a$& \textbf{0.150} & 0.748 & 0.401 & 0.531 & 0.263 & 0.583 & 0.207 & 0.707 & 0.250 & 0.424 & 0.232 & 0.466 \\
    Baseline&+ $\Lambda_x \cup \Lambda_v$ & 0.141 & \textbf{0.754} & \textbf{0.419} & \textbf{0.545} & \textbf{0.267} & \textbf{0.591}& \textbf{0.212} & \textbf{0.713} & \textbf{0.266} & \textbf{0.447} & \textbf{0.239} & 0.473 \\
  \hline
	\end{tabular}
	}
		\label{table:ablation_lm}
\end{table*}

To validate the effectiveness of our landmark selection strategy, we conducted several experiments with several ablative strategies on REVERIE~\cite{qi2020reverie} and R2R~\cite{anderson2018vision} \texttt{val} \texttt{unseen} splits. The results are shown in \cref{table:ablation_lm}.

\#1 is the baseline result without landmarks and the CoT process. The model in \#2 uses only landmarks from instructions $\Lambda_x$ in CoTL. Compared to \#1, the \texttt{SPICE} metric remarkably increases, which indicates a more accurate description of object relations in the instructions. Other metrics fluctuate. Based on \#2, the model in \#3 adds visual landmarks via spatial selection, which are denoted as $\Lambda_a$. Compared to \#2, almost all metrics rise, which demonstrates the value of visual landmarks. The model in \#4 adds visual landmarks via spatial and temporal selection $\Lambda_v$ in addition to landmarks from instructions, resulting in an increase in almost all scores compared to \#3. The results above further confirm the effectiveness of the proposed landmark selection mechanism.

\subsection{Values of $\beta$}
\label{sec:ablation_beta}

\begin{table}[tb]
\caption{Ablations on the value of $\beta$ in landmark selection (\S\ref{sec:ablation_beta}) on REVERIE~\cite{qi2020reverie} \texttt{val} \texttt{unseen} and R2R~\cite{anderson2018vision} \texttt{val} \texttt{unseen}.}
\centering
\resizebox{1\linewidth}{!}{
			\setlength\tabcolsep{2pt}
   \renewcommand\arraystretch{1.25}
\begin{tabular}{|c||cccccc|cccccc|} 
\hline\thickhline
\multirow{2}{*}{$\beta$} & \multicolumn{6}{c|}{REVERIE \texttt{val} \texttt{unseen}} & \multicolumn{6}{c|}{R2R \texttt{val} \texttt{unseen}}\\ 
\cline{2-13}
 & \multicolumn{1}{l}{\texttt{SPICE}~$\uparrow$} & \multicolumn{1}{l}{\texttt{BLEU-1}~$\uparrow$} &\multicolumn{1}{l}{\texttt{BLEU-4}~$\uparrow$} & \multicolumn{1}{l}{\texttt{CIDEr}~$\uparrow$} & \multicolumn{1}{l}{\texttt{Meteor}~$\uparrow$} & \multicolumn{1}{l|}{\texttt{Rouge}~$\uparrow$}
 & \multicolumn{1}{l}{\texttt{SPICE}~$\uparrow$} & \multicolumn{1}{l}{\texttt{BLEU-1}~$\uparrow$} &\multicolumn{1}{l}{\texttt{BLEU-4}~$\uparrow$} & \multicolumn{1}{l}{\texttt{CIDEr}~$\uparrow$} & \multicolumn{1}{l}{\texttt{Meteor}~$\uparrow$} & \multicolumn{1}{l|}{\texttt{Rouge}~$\uparrow$}  \\ 
\hline\hline

$0$ & \textbf{0.150} & 0.749 & 0.409 & 0.538 & 0.267  & 0.587 & 0.208 & \textbf{0.719} & 0.266 & 0.413 & 0.236  & 0.469 \\
$0.25$ & 0.141 & \textbf{0.754} & \textbf{0.419} & \textbf{0.545} & \textbf{0.267} & \textbf{0.591} & \textbf{0.212} & 0.713 & \textbf{0.266} & \textbf{0.447} & \textbf{0.239}  & \textbf{0.473} \\
$0.5$ & 0.137 & 0.717 & 0.376 & 0.488 & 0.262  & 0.576 & 0.206 & 0.692 & 0.247 & 0.410 & 0.231  & 0.461 \\

\hline
\end{tabular}}
		\label{table:ablation_beta}
\end{table}

We conduct ablations on the numerical values of $\beta$ in landmark selection on REVERIE~\cite{qi2020reverie} and R2R~\cite{anderson2018vision} \texttt{val} \texttt{unseen} splits. The results are presented in \cref{table:ablation_beta}. It can be observed that assigning $\beta$ to $0.25$ achieves the best performance.

\section{Further Analysis on STMT}
\label{sec:stmt_analysis}

\begin{figure}[tb]
    \centering
    \includegraphics[width=0.55\linewidth]{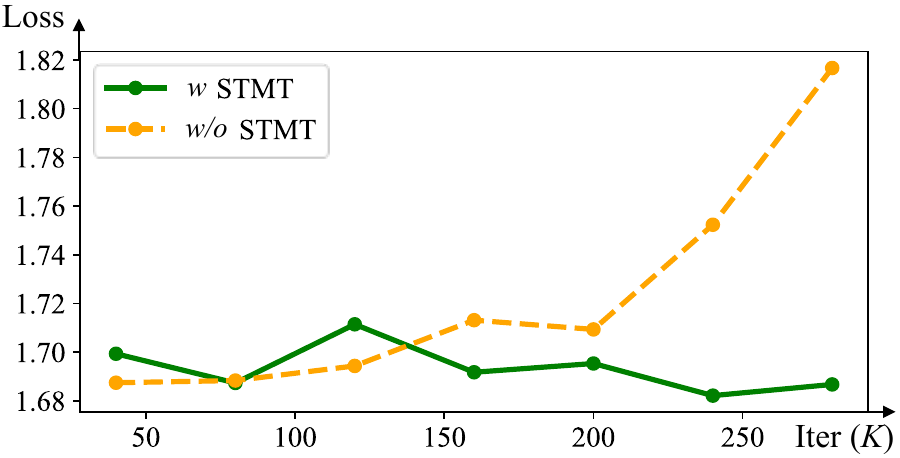}
      \caption{Validation loss on R2R~\cite{anderson2018vision} \texttt{val} \texttt{unseen} (\S\ref{sec:stmt_analysis}).}
   \label{fig:vis_loss}
\end{figure}

In \cref{fig:vis_loss}, we plot the curve illustrating the model's validation loss on R2R~\cite{anderson2018vision} \texttt{val} \texttt{unseen} during the training process. Compared to the baseline without STMT, We can observe that STMT effectively prevents overfitting as evidenced by the fact that its validation loss does not exhibit a gradual increase compared to the baseline. STMT effectively ensures the training stability of \textsc{C-Instructor} and also enhances the instruction quality.

\section{Additional Qualitative Results}
\label{sec:extra_vis}

\begin{figure*}[tb]
    \centering
    \includegraphics[width=\linewidth]{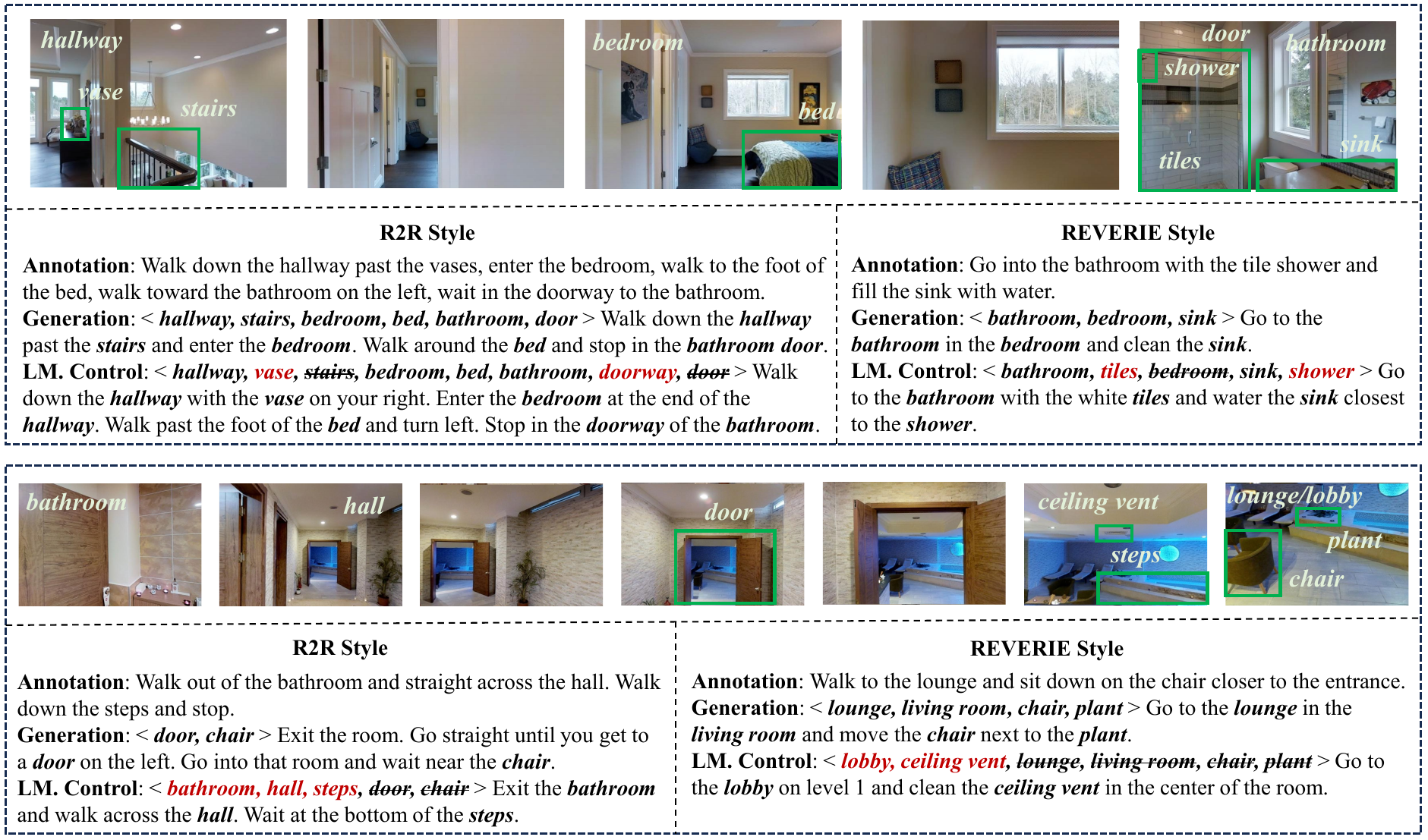}
      \caption{Additional visualizations of navigation trajectories and instruction generation results on R2R~\cite{anderson2018vision} and REVERIE~\cite{qi2020reverie} (\S\ref{sec:extra_vis}).}
   \label{fig:vis_supp}
\end{figure*}

In \cref{fig:vis_supp}, we provide more visualizations of navigation trajectories and corresponding instruction generation results. As observed, \textsc{C-Instructor} effectively identifies essential landmarks in the trajectory and generates high-quality instructions accordingly in specified linguistic styles. Control over the focus of \textsc{C-Instructor} can be achieved by manipulating landmarks. Modifying either part of landmarks (\cref{fig:vis_supp} upper) or all the landmarks (\cref{fig:vis_supp} lower) leads to reasonable instruction generation results.

\begin{figure*}[tb]
    \centering
    \includegraphics[width=\linewidth]{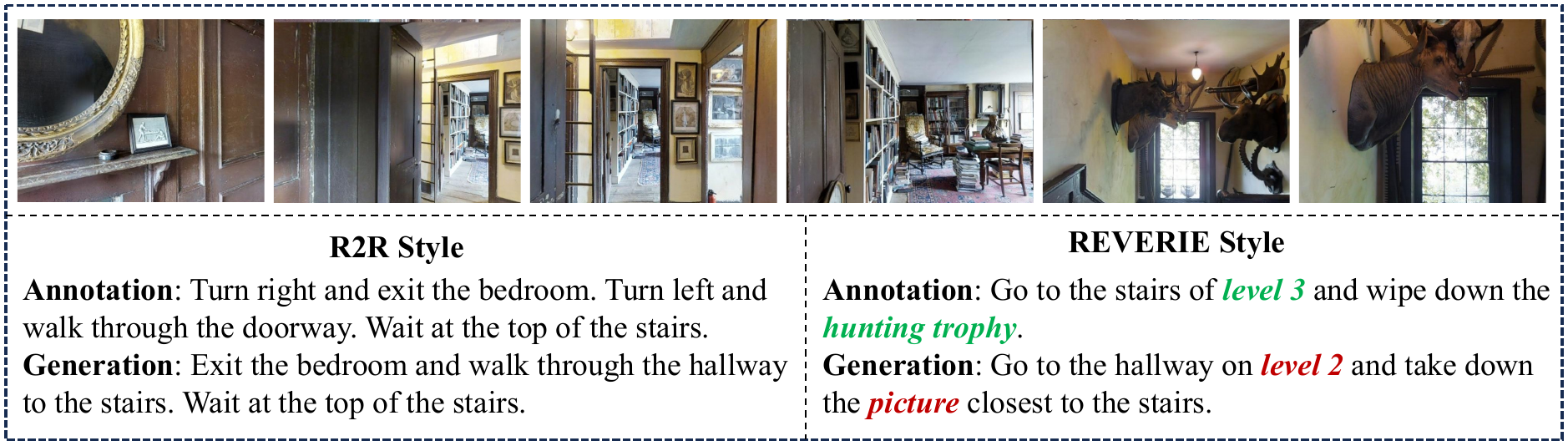}
      \caption{Failure case of \textsc{C-Instructor} (\S\ref{sec:extra_vis}).}
   \label{fig:vis_failure}
\end{figure*}

\noindent\textbf{Failure Case.} We present a failure case of \textsc{C-Instructor} in \cref{fig:vis_failure}. In this case, \textsc{C-Instructor} mistakes \textit{level 3} as \textit{level 2} for lack of knowledge of the global structure of the house. Furthermore, it misidentifies the rarely-seen object \textit{hunting trophy} as a \textit{picture}. This case suggests future efforts on global environmental structure encoding and more accurate object identification.

\section{More Discussion}
\label{sec:discussion}

\noindent\textbf{Social Impact.} \textsc{C-Instructor} can be used to provide feedback from intelligent embodied agents to humans as well as to guide humans who are unfamiliar with the environment. It can also serve as accessibility facilities for the visually impaired to find their way.

\noindent\textbf{Limitations.} Due to data availability, \textsc{C-Instructor} is trained on simulated data with discrete viewpoints, which limits its performance in real-world continuous environments. Moreover, as discussed in \S\ref{sec:extra_vis}, \textsc{C-Instructor} possesses limited ability in modeling the global structure of the environment, resulting in inaccurate instructions when referring to the global location of a specific object or room in the environment. 

\noindent\textbf{Future Work.} We plan to devise a mechanism that encodes the global structure of the environment into the instruction generator. With knowledge of the environment, the instruction generator can locate the user according to free-form natural language descriptions and provide path guidance according to the destination designated by the user.


\end{document}